\documentclass[runningheads]{llncs}

\usepackage[T1]{fontenc}
\usepackage{graphicx}
\usepackage{subfiles}
\usepackage{booktabs}
\usepackage{multirow}
\usepackage{caption}
\usepackage{amsmath}
\usepackage{amssymb}
\usepackage{arydshln}
\usepackage{marvosym}
\usepackage[table]{xcolor}
\usepackage{color}
\usepackage{url}
\usepackage{comment}
\usepackage{makecell}

\usepackage[colorlinks=true, linkcolor=blue, citecolor=blue, urlcolor=blue]{hyperref}

\begin{document}
\title{Morphology-Aware Channel-Decoupled Mamba Network for Medical Image Segmentation}
\titlerunning{Morphology-Aware Channel-Decoupled Mamba}
%

\author{Chenchen Wang\inst{1}}

\authorrunning{Wang}

\institute{
$^1$School of Computer Science and Informatics, Cardiff University, Cardiff, UK
}

\maketitle           
\begin{abstract}
Accurate multi-organ segmentation in computed tomography (CT) is essential for computer-aided diagnosis, treatment planning, and disease monitoring. However, substantial variations in organ shape and size, together with low tissue contrast and ambiguous boundaries, complicate the modeling of heterogeneous anatomical structures. To address these challenges, we propose a Morphology-Aware Channel-Decoupled Mamba Network (MCDNet), which integrates adaptive local structural enhancement with channel-decoupled long-range modeling. Specifically, we introduce a Morphology-Aware Channel-Decoupled State-Space (MCSS) block that divides feature channels into separate processing groups, allowing differentiated structural representations to be learned within an efficient state-space framework. Within each group, a Morphology-Aware Enhancement Module (MAEM) combines deformable convolutions with channel attention to adapt local feature extraction to anatomical shape variations. Group-wise state-space modeling then captures long-range spatial dependencies, and channel gating adaptively integrates the resulting features. Experiments on the Synapse abdominal CT dataset demonstrate the effectiveness of MCDNet, achieving an average Dice Similarity Coefficient of 81.61\% for multi-organ segmentation.

\end{abstract}

\section{Introduction}
\label{sec:intro}
Medical image segmentation plays a vital role in clinical workflows by enabling automated delineation of organs and pathological regions to support diagnosis, treatment planning, and disease monitoring. In abdominal computed tomography (CT), substantial variations in organ shape and size, low soft-tissue contrast, and similar intensity distributions between adjacent organs obscure anatomical boundaries and complicate structure identification. These challenges require segmentation methods to capture both the global spatial context of multiple organs and the local morphological details of individual structures.

Existing approaches to medical image segmentation rely predominantly on deep learning. Convolutional neural networks with U-shaped encoder--decoder architectures, such as UNet\cite{unet}, nnUNet\cite{nnunet}, and UNet++\cite{UNet++}, extract fine-grained features through multi-level skip connections and have become the backbone of many systems. More recently, Transformer-based models, such as Swin-Unet\cite{swinunet}, TransUNet\cite{transunet}, and nnFormer\cite{nnformer}, apply self-attention mechanisms to capture long-range dependencies and global context. However, the quadratic computational complexity of self-attention with respect to the number of image tokens imposes a significant processing burden. State space models\cite{mamba} address this challenge by enabling efficient long-range modeling with linear complexity, motivating a series of segmentation methods\cite{icassp1,icassp2}, including VM-UNet\cite{vmunet} and Swin-UMamba\cite{swinumamba}.

Despite these advances, efficient long-range modeling alone does not explicitly account for the morphological diversity of abdominal organs. Structures with different shapes and spatial extents require different local representations, while subtle contours between adjacent organs are difficult to resolve using global context alone. Processing all feature channels through a shared state-space transformation may also limit the capacity to represent these heterogeneous structural patterns separately. These observations motivate a joint design that adapts local feature extraction to organ morphology and decouples channel processing to support differentiated structural representations.

To address these challenges, we introduce MCDNet, which integrates morphology-aware enhancement and channel-decoupled state-space modeling within a U-shaped encoder--decoder. Our Morphology-Aware Channel-Decoupled State-Space (MCSS) block divides feature channels into separate processing groups. In each group, a Morphology-Aware Enhancement Module (MAEM) adjusts convolutional sampling locations and reweights channel responses to capture local shape variations, followed by a State-Space 2D (SS2D) unit for long-range spatial modeling. An adaptive channel gate integrates the group outputs. This design connects local morphological adaptation with efficient global context modeling while allowing different channel groups to learn complementary structural features. Our contributions are as follows:
\begin{itemize}
\item[$\bullet$] We propose MCDNet, which integrates morphology-aware local enhancement with channel-decoupled state-space modeling for medical image segmentation.
\item[$\bullet$] We design an MCSS block that processes channel groups separately and adaptively integrates their outputs, together with an MAEM that uses deformable sampling and channel attention to capture diverse anatomical morphologies.
\item[$\bullet$] Experiments on the Synapse abdominal CT dataset demonstrate the effectiveness of MCDNet, which achieves the highest average DSC among the compared methods.
\end{itemize}

\begin{figure*}[t]
\centerline{\includegraphics[width=\textwidth]{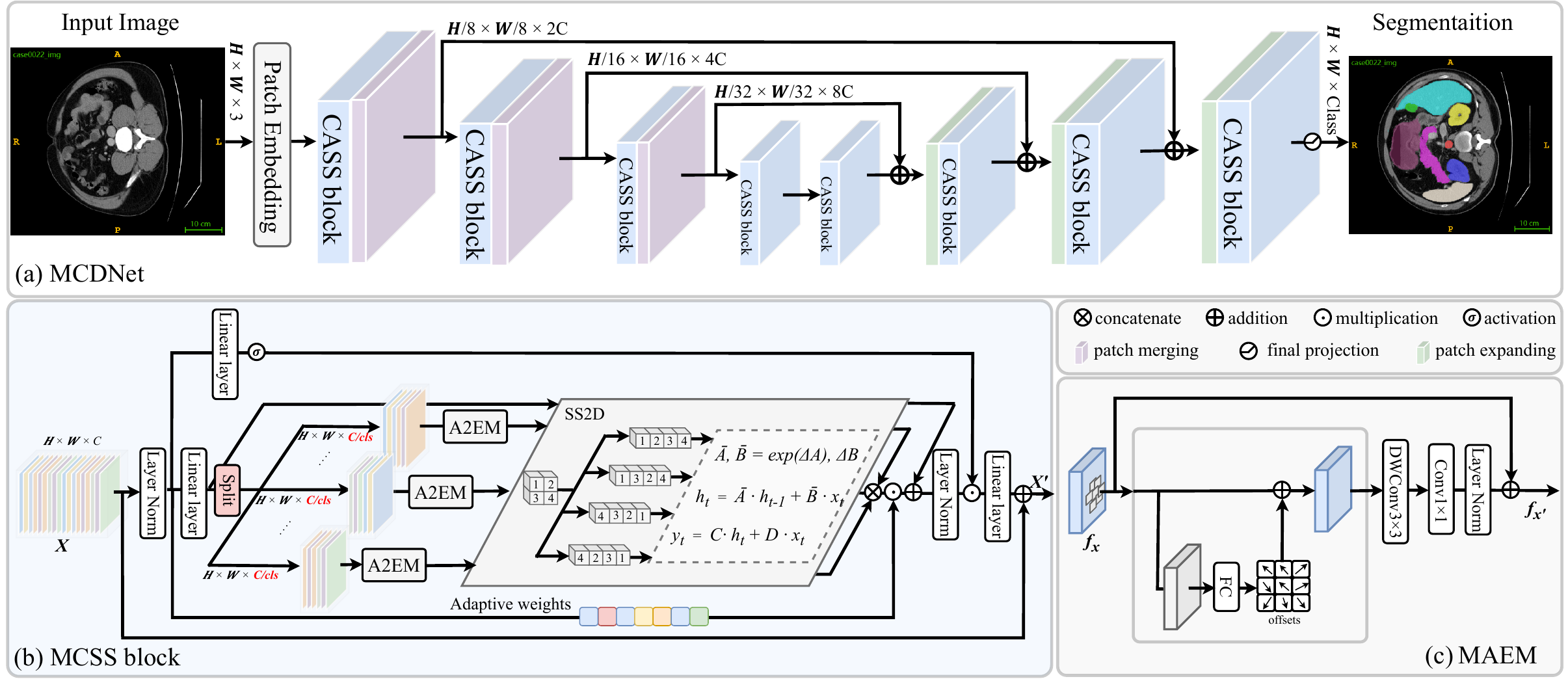}}
\caption{Overview of MCDNet. (a) The U-shaped encoder--decoder is built from Morphology-Aware Channel-Decoupled State-Space (MCSS) blocks. (b) Each MCSS block combines group-wise morphology-aware enhancement, SS2D-based long-range modeling, and adaptive channel gating. The Morphology-Aware Enhancement Module (MAEM) integrates deformable convolutions, depthwise--pointwise convolutions, and channel attention.}  
\label{fig1}
\end{figure*}
\section{Method}
\subsection{Overview}
As illustrated in Figure~\ref{fig1}-(a), MCDNet first embeds the input CT slice into patches and processes the resulting features through a U-shaped encoder--decoder built from MCSS blocks. Each block combines morphology-aware local enhancement with channel-decoupled long-range modeling. A final class projection produces a segmentation tensor of size $H\times W\times C$, where $C$ denotes the number of anatomical classes.
\subsection{Morphology-Aware Channel-Decoupled State-Space Block}

The MCSS block couples adaptive local feature extraction with separate state-space processing of channel groups. As shown in Figure~\ref{fig1}-(b), it incorporates an MAEM into each group before SS2D-based spatial modeling, allowing local shape information and long-range context to be modeled within the same processing stream. Given an input feature map $X\in\mathbb{R}^{H\times W\times C}$, where $C$ denotes the feature-channel dimension in this block, we first apply LayerNorm and a linear projection, and then split the channels into $N$ groups. The group count is set to the number of segmentation classes, providing an anatomical basis for the channel decomposition. The resulting groups provide separate feature streams in which differentiated structural representations are learned end to end through segmentation supervision. For equal group widths, the split is written as
\begin{align}
F^{(i)} = Split(X), F^{(i)}\in\mathbb{R}^{H\times W\times \tfrac{C}{N}},\quad i=1,\dots,N.
\end{align}

When $C$ is not divisible by $N$, the remaining channels are assigned to the last group. The channel partition is fixed, while the feature transformations and fusion weights are learned. Each group is first refined by an MAEM to adapt its local responses to morphological variation, and then independently processed by an SS2D unit to capture long-range spatial dependencies. Let $x_t$ denote a feature at spatial position $t$; learnable increments $\Delta A,\Delta B\in\mathbb{R}^{d\times d}$ generate transition matrices for the state-space recurrence:
\begin{align}
\bar{A} &= \exp(\Delta A),  & \bar{B} &= \Delta B,\\
h_t &= \bar{A}\,h_{t-1} + \bar{B}\,x_t, & y_t &= C\,h_t + D\,x_t,
\end{align}

To adaptively integrate the group-wise outputs, we compute a gate for each channel using channel attention. Denote the residual input as $X_{res}\in\mathbb{R}^{B\times H\times W\times d}$. We normalize with LayerNorm and aggregate by global average pooling:
\begin{align}
Z_c = \frac{1}{H\,W}\sum_{h=1}^{H}\sum_{w=1}^{W}\mathrm{LayerNorm}(X_{res}) , 
\quad Z_c\in\mathbb{R}^{B\times d}.
\end{align}
A two-layer MLP with reduction ratio $r$ and ReLU, followed by Sigmoid, produces gating weights:
\begin{align}
G = \sigma\bigl(W_2\,\mathrm{ReLU}(W_1\,Z_c)\bigr), 
\quad G\in\mathbb{R}^{B\times d}.
\end{align}
We reshape \(G\) to \(G' \in \mathbb{R}^{B\times 1\times 1\times d}\), spatially broadcast it, and concatenate the SS2D outputs along the channel dimension. The concatenated features are weighted by \(G'\), allowing the block to adjust the contributions of different channel responses. LayerNorm and a linear projection then restore $\widetilde{F}\in\mathbb{R}^{H\times W\times C}$, and a residual addition with $X$ yields the block output. Stacking MCSS blocks in the encoder--decoder connects morphology-aware local enhancement with channel-decoupled long-range modeling throughout MCDNet.  
\subsection{Morphology-Aware Enhancement Module}
To explicitly model local morphological variation, we introduce an MAEM within each channel group. Organs vary in shape, scale, and boundary geometry, making a fixed sampling pattern potentially insufficient for their local representation. MAEM predicts deformable convolutional offsets from the input features to adapt sampling locations to local image structure. Depthwise--pointwise convolutions then refine the sampled responses, and channel attention selectively weights the resulting features. These operations are combined within a residual framework to preserve the input representation. Given an input feature map $X\!\in\!\mathbb{R}^{B\times C\times H\times W}$, the local feature transformation is defined as:
\begin{align}
\Delta         &= \operatorname{Conv}_{\text{offset}}(X),Y_{\mathrm{dcn}} = \operatorname{DCN}(X,\Delta) \\
Y_{\mathrm{dw}}  &= \operatorname{PW}\!\bigl(\operatorname{DW}(Y_{\mathrm{dcn}})\bigr),\widehat{Y} = \operatorname{SiLU}\!\bigl(\operatorname{BN}(Y_{\mathrm{dw}})\bigr)
\end{align}
where $\Delta\in\mathbb{R}^{B\times 2K^2\times H\times W}$ denotes the learned offsets for a deformable convolution with a $K\times K$ sampling grid. DCN($\cdot$) is the deformable convolution, DW($\cdot$) and PW($\cdot$) denote depthwise and pointwise convolutions, respectively, BN denotes batch normalization, and SiLU is the activation. Next, MAEM applies lightweight channel attention to reweight the enhanced features:
\begin{align}
    z &= \mathrm{GAP}(\widehat{Y})\in\mathbb{R}^{B\times C\times 1\times 1}\,, \\
    s &= \sigma\!(W_2(\mathrm{ReLU}(W_1(z)))), Y' = s \,\odot\, \widehat{Y}\,,
\end{align}
where GAP is global average pooling, $W_1\!\in\!\mathbb{R}^{\frac{C}{r}\times C}$ and $W_2\!\in\!\mathbb{R}^{C\times\frac{C}{r}}$ are pointwise conv layers with reduction ratio $r$, $\sigma$ is the sigmoid, and $\odot$ denotes channel-wise multiplication. Finally, a residual connection adds the enhanced features to the input representation.

Deformable sampling provides spatial adaptation to local shape variation, while channel attention adjusts the importance of the extracted responses. Within an MCSS block, MAEM therefore provides morphology-aware local features for the subsequent group-wise state-space transformation, complementing its long-range spatial modeling.  
\subsection{End-to-End Learning}
We train the entire network in an end-to-end manner using the Dice loss. Given a predicted segmentation mask \(\hat{Y}\) and the ground-truth mask \(Y\), we minimize
\[
\mathcal{L}_{\mathrm{Dice}}
= 1 - \frac{2 \sum_{i} \hat{Y}_i \, Y_i + \epsilon}
{\sum_{i} \hat{Y}_i + \sum_{i} Y_i + \epsilon}\,,
\]
where the summations run over all pixel locations and \(\epsilon\) is a small constant for numerical stability. 

\section{Experiments}
\subsection{Dataset}
We evaluate our approach on the Synapse Multi-Organ Segmentation dataset\cite{synapse}, a publicly available collection of 30 abdominal CT cases comprising 3,779 axial CT slices. We report segmentation performance for eight anatomical structures: the aorta, gallbladder, left kidney, right kidney, liver, pancreas, spleen, and stomach.

\subsection{Evaluation Metrics}
We evaluate segmentation performance using the Dice Similarity Coefficient (DSC) and the 95\% Hausdorff Distance (HD95). DSC measures the overlap between predicted and ground-truth segmentations, while HD95 assesses boundary discrepancy. Higher DSC and lower HD95 indicate better performance.

\subsection{Implementation Details}
We trained our model on a GeForce RTX 4090 GPU for 300 epochs. CT slices were resized to $224\times224$, with a batch size of 22 and an initial learning rate of $1\times10^{-3}$.

\begin{table*}[t]
    \centering
    \caption{Comparison with state-of-the-art methods on the Synapse\cite{synapse} dataset. Average DSC (\%), HD95, and per-organ DSC (\%) are reported.}
        \resizebox{1\linewidth}{!}{
    \begin{tabular}{l|cc|cccccccc}
       \midrule
         \textbf{Model}&
         \textbf{DSC$\uparrow$}&
         \textbf{HD95}$\downarrow$ &
         \textbf{Aorta} &
         \textbf{Gallbladder} &
         \textbf{Kidney(L)} &
         \textbf{Kidney(R)} &
          \textbf{Liver}&
          \textbf{Pancreas}&
          \textbf{Spleen}&
          \textbf{Stomach} \\ \hline 
      
          MedSAM\cite{medsam}&
          73.99&
          23.68&
          67.15&
          60.99&
          67.68&
          66.79&
          88.92&
          70.72&
          87.20&
          82.48\\

          U-Net\cite{unet}&
          76.85&
          39.70&
          89.07&
          \textbf{69.72}&
          77.77&
          68.60&
          93.43&
          53.98 &
          86.67&
          75.58\\
          Att-UNet\cite{attentionunet}&
          77.77&
          36.02&
          89.55&
          68.88&
          77.98&
          71.11&
          93.57&
          58.04&
          87.30&
          75.75\\
     
          TransUNet\cite{transunet}&
          77.48&
          31.69&
          87.23&
          63.13&
          81.87&
          77.02&
          94.08&
          55.86&
           85.08 &
          75.62\\
       
          TransDeepLab\cite{transdeeplab}&
          80.16&
          21.25&
          86.04&
          69.16&
          84.08&
          79.88&
          93.53&
          61.19&
          89.00&
          78.40\\

          MEW-UNet\cite{mewunt}&
          78.92&
          \textbf{16.44}&
          86.68&
          65.32&
          82.87&
          80.02&
          93.63&
          58.36&
          90.19&
          74.26\\
          SAM3D\cite{sam3d}&
          79.56&
          17.87&
           85.64&
          86.31&
         84.29&
          49.81 &
        76.11&
        \textbf{69.32}&
          89.57&
          95.42\\

          MedMamba\cite{medmamba}&
          80.67&
          28.15&
          \textbf{88.23}&
          \textbf{69.53}&
          82.22&
          79.49&
          \textbf{96.89}&
          58.57&
          92.29&
          79.12\\          
          VM-UNet\cite{vmunet}&
          81.08&
          19.21&
          86.40&
          69.41&
          86.16&
          \textbf{82.76}&
          94.17&
          58.80&
          89.51&
          \textbf{81.40}\\
          \midrule
           \textbf{MCDNet (ours)}&
          \textbf{81.61}&
          23.72&
          86.99&
         69.70&
           85.48 &
          82.45&
          94.39&
          63.57&
          \textbf{90.90}&
          79.39
          \\\midrule

    \end{tabular}
            }
    \label{tab:synapse}
 
\end{table*}

\subsection{Comparison with State-of-the-Art Methods}
We compare MCDNet with recent state-of-the-art medical image segmentation methods on the Synapse abdominal CT dataset. Quantitative results are reported in Table~\ref{tab:synapse}. MCDNet achieves the highest average DSC of 81.61\% among the compared methods, improving upon VM-UNet and MedMamba by 0.53 and 0.94 percentage points, respectively. Compared with VM-UNet, MCDNet increases the DSC for the pancreas from 58.80\% to 63.57\% and for the spleen from 89.51\% to 90.90\%. These results demonstrate the performance of MCDNet across organs with diverse morphologies. Its design combines local morphological adaptation with channel-decoupled long-range modeling. MCDNet achieves an HD95 of 23.72, although several compared methods obtain lower HD95 values.

\section{Conclusion}
We propose MCDNet, a Morphology-Aware Channel-Decoupled Mamba Network for medical image segmentation. Its MCSS block processes channel groups separately to learn differentiated structural representations and captures long-range spatial context through SS2D. Within each group, MAEM uses deformable sampling and channel attention to adapt local feature extraction to anatomical shape variations. Adaptive channel gating integrates the group outputs, connecting morphology-aware enhancement with channel-decoupled state-space modeling. Experiments on the Synapse abdominal CT dataset demonstrate improved average segmentation accuracy over the compared methods, achieving an average DSC of 81.61\%.

\bibliographystyle{splncs04}
\bibliography{Paper-0313}
\end{document}